%% file: main.tex
\documentclass[runningheads]{llncs}
\usepackage[T1]{fontenc}
\usepackage{graphicx}
\usepackage{booktabs}
\usepackage[misc]{ifsym}
\newcommand{\corr}{(\Letter)}

\input{src/acronyms}

\usepackage{hyperref}
\usepackage{url}

\usepackage{amssymb}
\usepackage{amsmath}
\usepackage{caption}
\usepackage{lipsum}
\usepackage{multirow}
\usepackage{multicol}
\usepackage{indentfirst}
\usepackage{float}
\usepackage{hyperref}
\usepackage{footnote}
\usepackage{fancyvrb}
\usepackage{algorithmic}
\usepackage[ruled,linesnumbered]{algorithm2e}
\usepackage{amsfonts}
\usepackage{siunitx}
\usepackage{color}
\usepackage{makecell}
\usepackage{xspace}
\usepackage{subcaption}
\usepackage{comment}
\usepackage{soul}
\usepackage{wrapfig}
\usepackage{enumitem}
\usepackage{booktabs}
\def \sys {{\rm FedTDD}\xspace}

\begin{document}

\title{Federated Time Series Generation on Feature and Temporally Misaligned Data}

\titlerunning{\sys}

\author{Zhi Wen Soi\inst{1} 
\and
Chenrui Fan\inst{1}\thanks{Equal contribution.}
\and
Aditya Shankar\inst{2}$^\star$
\and
Abele M\u{a}lan\inst{3} \and
Lydia Y. Chen\inst{2,3} \corr}
\authorrunning{Z.W. Soi et al.}

\tocauthor{Zhi Wen Soi, Chenrui Fan, Aditya Shankar, Abele M\u{a}lan, Lydia Y. Chen}
\toctitle{Federated Time Series Generation on Feature and Temporally Misaligned Data}

\institute{University of Bern, Switzerland \email{\{zhi.soi,chenrui.fan\}@students.unibe.ch}
\and
TU Delft, Netherlands \email{a.shankar@tudelft.nl,lydiachen@ieee.org}
\and
University of Neuch\^atel, Switzerland
\email{abele.malan@unine.ch}}

\maketitle

\begin{abstract}

Distributed time series data presents a challenge for federated learning, as clients often possess different feature sets and have misaligned time steps. Existing federated time series models are limited by the assumption of perfect temporal or feature alignment across clients. In this paper, we propose \sys, a novel federated time series diffusion model that jointly learns a synthesizer across clients. At the core of \sys is a novel data distillation and aggregation framework that reconciles the differences between clients by imputing the misaligned timesteps and features. In contrast to traditional federated learning, \sys learns the correlation across clients' time series through the exchange of local synthetic outputs instead of model parameters. A coordinator iteratively improves a global distiller network by leveraging shared knowledge from clients through the exchange of synthetic data. As the distiller becomes more refined over time, it subsequently enhances the quality of the clients' local feature estimates, allowing each client to then improve its local imputations for missing data using the latest, more accurate distiller.  Experimental results on five datasets demonstrate \sys 's effectiveness compared to centralized training, and the effectiveness of sharing synthetic outputs to transfer knowledge of local time series. Notably, \sys achieves 79.4\% and 62.8\% improvement over local training in Context-FID and Correlational scores. Our code is available at: \url{https://github.com/soizhiwen/FedTDD}.


\end{abstract}

\input{src/introduction}
\input{src/related_work}
\input{src/methodology}
\input{src/experiments}
\input{src/conclusion}
\input{src/bibliography}

\end{document}

%% file: src/acronyms.tex
\usepackage{acronym}

\acrodef{pr}[$\mathbf{PR}$]{public ratio}
\acrodef{sr}[$\mathbf{SR}$]{split ratio}
\acrodef{mr}[$\mathbf{MR}$]{missing ratio}
\acrodef{pisalpha}[$\mathbf{\alpha}$]{PIS alpha}

%% file: src/introduction.tex
\section{Introduction}

\begin{figure}
    \centering
    \includegraphics[width=\linewidth]{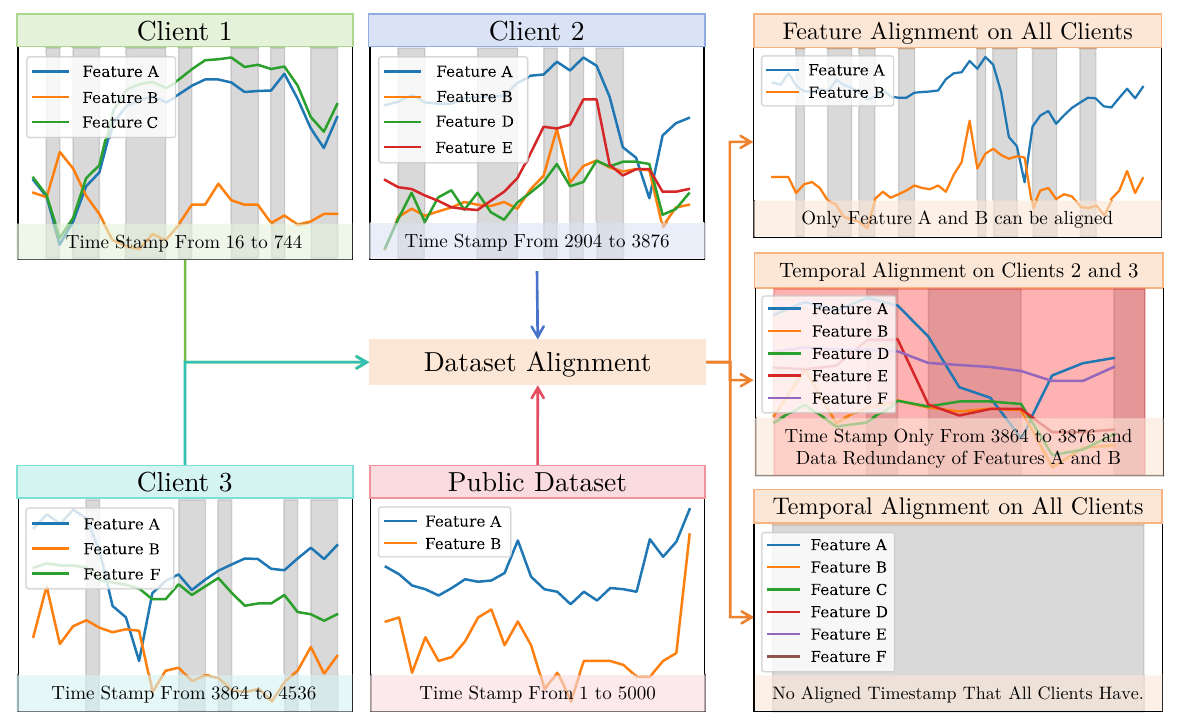}
    \caption{Feature and temporally misaligned time series. The grey masking indicates missing data.}
    \label{fig:problem}
\end{figure}

Multivariate time series data are pivotal in many domains, such as healthcare, finance, manufacturing, and sales~\cite{lim2021time}. Consider a collaboration between multiple clients, shown in \autoref{fig:problem}. In a healthcare setting, these clients could be hospitals, each collecting patient data locally for a downstream task, such as predicting patient outcomes. The data gathered, such as vital signs like heart rate and blood pressure, is inherently \textit{temporal}, i.e., time series data. Aggregating data from all the sources could improve model performance due to increased sampled diversity when training downstream predictive models. However, privacy regulations such as the General Data Protection Regulation (GDPR) and confidentiality agreements between hospitals prevent sharing of raw data~\cite{gdpr,alaa2021generative,meijer2025ts}.

\textit{Federated learning} (FL)~\cite{mcmahan2017communication} takes a step towards tackling this privacy challenge by enabling clients to train a global model by sharing locally trained model parameters rather than raw data. However, this environment faces the challenge of \textit{feature and temporal misalignment}~\cite{luu2021time}, as hospitals may possess different feature sets with varying time intervals for data collection.

In \textit{horizontal} FL~\cite{li2020federated}, different clients have data for the same features but for different samples or timesteps. Hence, it can tackle situations involving temporal misalignment but not feature misalignment. On the other hand, in \textit{vertical} FL~\cite{liu2024vertical}, different clients possess different feature sets for the same samples or timesteps. While this can handle feature misalignment, it cannot tackle temporal misalignment. Hence, neither horizontal nor vertical FL can fully tackle scenarios with both feature and temporal misalignment. On top of this, data may be missing or incomplete due to unavailability or inconsistent collection frequencies, further hindering a model's ability to learn patterns~\cite{pratama2016review}.

To overcome these limitations, we propose \textbf{FedTDD} (\underline{Fed}erated Learning in Multivariate \underline{T}ime Series via \underline{D}ata \underline{D}istillation), a first-of-its-kind federated time series diffusion model capable of learning a time series synthesizer from clients' distinct features with temporal misalignment. \sys introduces a novel data distillation~\cite{sachdeva2023data} and aggregation framework for the common feature set, whose values differ across clients and can be obtained from the public domain. In this framework, a coordinator maintains a global model called the \emph{distiller}, trained iteratively using a combination of public data and clients' intermediate synthetic data outputs. Each client keeps a local time series diffusion model for imputing local features which leverages the latest distiller to improve the quality of local estimates. Unlike traditional federated learning, \sys learns the correlations among clients' time series through the exchange of synthetic outputs instead of aggregating models~\cite{mcmahan2017communication}, effectively handling feature and temporal misalignment without sharing raw data. 

Given the recent advancements of diffusion models over mainstream generative models like \textit{Generative Adversarial Networks} (GANs)~\cite{goodfellow2020generative}, we utilize a time series \textit{Denoising Diffusion Probabilistic Model} (DDPM)~\cite{ho2020denoising}, adapted to handle temporal dependencies through temporal embeddings and sequential conditioning. Specifically, we select \textbf{Diffusion-TS}~\cite{yuan2024diffusion} since it leverages both time and frequency domain information, effectively capturing trends and seasonality, which leads to a more accurate imputation of missing data. By imputing data from unaligned time steps, clients can obtain temporally aligned data without needing alignment on the features or sharing raw data.

In summary, our major contributions are as follows: 
(i) We propose a novel federated generative learning framework that effectively handles temporal and feature-level misalignment and data missing problems in time series data.
(ii) We develop a data distillation and aggregation framework that learns correlations among clients' time series by exchanging synthetic data instead of model parameters, enabling clients to improve their local models without direct data sharing and effectively handling data discrepancies.
(iii) We conduct extensive experiments on five benchmark datasets, showing up to 79.4\% and 62.8\% improvement over local training in Context-FID and Correlational scores under extreme feature and temporal misalignment cases and achieving performance comparable to centralized training.

%% file: src/related_work.tex
\section{Related Work}

\input{src/tables/related_work}

\paragraph{Time series generation} Generative models for time series data aim to capture temporal dependencies and sequential patterns inherent in such datasets. TimeGAN~\cite{yoon2019time} combines \textit{generative adversarial networks} (GANs)~\cite{goodfellow2020generative} with recurrent neural networks~\cite{mogren2016c} to produce realistic multivariate time series. TimeVAE~\cite{desai2021timevae} utilizes variational autoencoders (VAEs)~\cite{kingma2013auto} tailored for time series to capture trends and seasonality. Recently, diffusion-based models like TimeGrad~\cite{rasul2021autoregressive}, CSDI~\cite{tashiro2021csdi}, SSSD~\cite{alcaraz2022diffusion}, TSDiff~\cite{kollovieh2024predict}, and Diffusion-TS~\cite{yuan2024diffusion} have further advanced time series generation by producing high-fidelity sequences, outperforming the mainstream GANs and VAE-based techniques. Despite their effectiveness, these models operate in centralized settings and assume fully aligned data with consistent features and timestamps. They are not equipped to handle feature and temporal misalignments common in real-world distributed scenarios, making them unsuitable for federated environments with heterogeneous data distributions~\cite{mendieta2022local,qu2022rethinking}.

\paragraph{Federated learning with generative models} Federated learning~\cite{zhang2021survey} has primarily been applied to image generation, such as FedCycleGAN~\cite{song2021federated} leverages CycleGAN~\cite{zhu2017unpaired} in federated settings to generate synthetic images while preserving data privacy. For tabular data, methods like GTV~\cite{zhao2023gtv}, DPGDAN~\cite{wang2023differentially}, and SiloFuse~\cite{shankar2024silofuse} employ GANs and diffusion models within vertical federated learning frameworks to synthesize tabular datasets. However, these approaches focus on vertically partitioned data, where all clients have features corresponding to the same sample ID, and do not address data redundancy or misalignment issues. Federated learning with generative models for time series data remains under-explored. Existing works such as FedGAN~\cite{rasouli2020fedgan}, VFLGAN-TS~\cite{yuan2024vflgan}, and T2TGAN~\cite{brophy2021estimation} extend GANs to federated time series generation. VFLGAN-TS operates in a vertical federated learning context, tackling feature misalignment, but does not handle temporal misalignment. In contrast, T2TGAN tackles horizontal federated learning settings but introduces data redundancy due to overlapping data among clients and cannot handle feature mismatches between clients. As summarized in \autoref{tab:related_work}, these methods encounter issues as shown in \autoref{fig:problem}, making them less effective for federated time series generation where both feature and temporal misalignments are prevalent.

\paragraph{Preliminary on generative modeling with DDPMs}

For the generative backbone, we adopt the Diffusion-TS architecture~\cite{yuan2024diffusion}, which extends DDPMs~\cite{ho2020denoising} to capture temporal patterns using a generative modeling process. DDPMs are models trained using a \textit{forward noising} and \textit{backward denoising} process. The forward phase progressively adds random Gaussian noise to the data $\mathbf{s}_0$ at diffusion step $t$, where the transition is parameterized by $q(\mathbf{s}_t \mid \mathbf{s}_{t-1}) = \mathcal{N}(\mathbf{s}_t; \sqrt{1 - \beta_t} \, \mathbf{s}_{t-1}, \beta_t \, \mathbf{I})$ with $\beta_t \in (0, 1)$, eventually transforming it into pure noise $\mathbf{s}_T \sim \mathcal{N}(0, \mathbf{I})$.
The backward phase is where the model learns to reverse this noising process. Starting from random noise $\mathbf{s}_T \sim \mathcal{N}(0, \mathbf{I})$, it iteratively removes the added noise step by step via $p_\theta(\mathbf{s}_{t-1} \mid \mathbf{s}_t) = \mathcal{N}(\mathbf{s}_{t-1}; \boldsymbol{\mu}_\theta(\mathbf{s}_t, t), \Sigma_\theta(\mathbf{s}_t, t))$, to reconstruct a new data sample resembling the original input distribution. The functions $\boldsymbol{\mu}_\theta$ and $\Sigma_\theta$ are generally estimated using a model.

Diffusion-TS extends standard DDPMs by incorporating mechanisms specifically designed for time series characteristics such as trends and seasonality~\cite{kitagawa1984smoothness}. Instead of treating data points independently, it utilizes an encoder-decoder transformer architecture~\cite{vaswani2017attention} that processes entire sequences, effectively modeling temporal relationships. To handle trends, Diffusion-TS decomposes the time series into components that represent slow-varying behaviors over time. For capturing seasonality and periodic patterns, it employs frequency domain analysis using the Fast Fourier Transform (FFT)~\cite{heckbert1995fourier}. By integrating FFT, the model can analyze and reconstruct cyclical patterns~\cite{ceneda2018guided} within the data, allowing it to learn both time and frequency domain representations~\cite{fons2022hypertime}. This combination enables Diffusion-TS to generate more accurate and realistic time series data by effectively modeling complex temporal dynamics. Besides, Diffusion-TS supports both \textit{unconditional} and \textit{conditional} generation. In the unconditional generation, the model produces new samples solely based on the learned data distribution, starting from random noise and applying the learned denoising process. In the conditional generation, Diffusion-TS utilizes gradient-based guidance during sampling to incorporate the observed data \( \mathbf{y} \). At each diffusion step, the model refines its estimated time series \( \hat{\mathbf{s}}_0 \) by adjusting it with a gradient term that enforces consistency with the observed data. The refinement can be computed via $\tilde{\mathbf{s}}_0(\mathbf{s}_t, t; \theta) = \hat{\mathbf{s}}_0(\mathbf{s}_t, t; \theta) + \eta \nabla_{\mathbf{s}_t} ( \left\| \mathbf{y} - \hat{\mathbf{s}}_0(\mathbf{s}_t, t; \theta) \right\|^2 + \gamma \log p(\mathbf{s}_{t-1} \mid \mathbf{s}_t) )$, where \( \eta \) is a hyperparameter that controls the strength of the gradient guidance, and \( \gamma \) balances the trade-off between fitting the observed data and maintaining the generative model's prior distribution \( p(\mathbf{s}_{t-1} \mid \mathbf{s}_t) \). This iterative refinement ensures that the generated time series aligns with the provided observations and preserves the temporal patterns learned during training. Further details of Diffusion-TS are shown in Appendix B.2.

%% file: src/tables/related_work.tex
\begin{table}
    \centering
    \caption{Overview of the related work.}
    \label{tab:related_work}
    \begin{tabular}{l c c c c c}
        \toprule[1.5pt]
          \multirow{2}{*}{Method}
        & \multirow{2}{*}{\makecell{Model \\ Type}}
        & \multirow{2}{*}{\makecell{Time \\ Series}}
        & \multirow{2}{*}{\makecell{FL \\ Type}}
        & \multirow{2}{*}{\makecell{Handles Temporal \\ Misalignment}} 
        & \multirow{2}{*}{\makecell{Handles Feature \\ Misalignment }} \\
        \\
        \midrule

          GTV~\cite{zhao2023gtv}               & GAN         &\texttimes       & Vertical     & \texttimes  & \checkmark        \\
          DPGDAN~\cite{wang2023differentially}  & GAN         &\texttimes       & Vertical     & \texttimes  & \checkmark        \\
          SiloFuse~\cite{shankar2024silofuse}   & DDPM        &\texttimes       & Vertical     & \texttimes  & \checkmark        \\
          VFLGAN-TS~\cite{yuan2024vflgan}       & GAN         &\checkmark   & Vertical     & \texttimes  & \checkmark        \\
          FedGAN~\cite{rasouli2020fedgan}       & GAN         &\checkmark   & Horizontal   & \checkmark  & \texttimes    \\
          T2TGAN~\cite{brophy2021estimation}    & GAN         &\checkmark   & Horizontal   & \checkmark  & \texttimes    \\
          \midrule
          \sys(Ours)                            & DDPM        &\checkmark   & Hybrid       & \checkmark  & \checkmark \\
        \bottomrule[1.5pt]
    \end{tabular}
\end{table}

%% file: src/methodology.tex
\section{FedTDD}

\begin{figure}
    \centering
    \includegraphics[width=\linewidth]{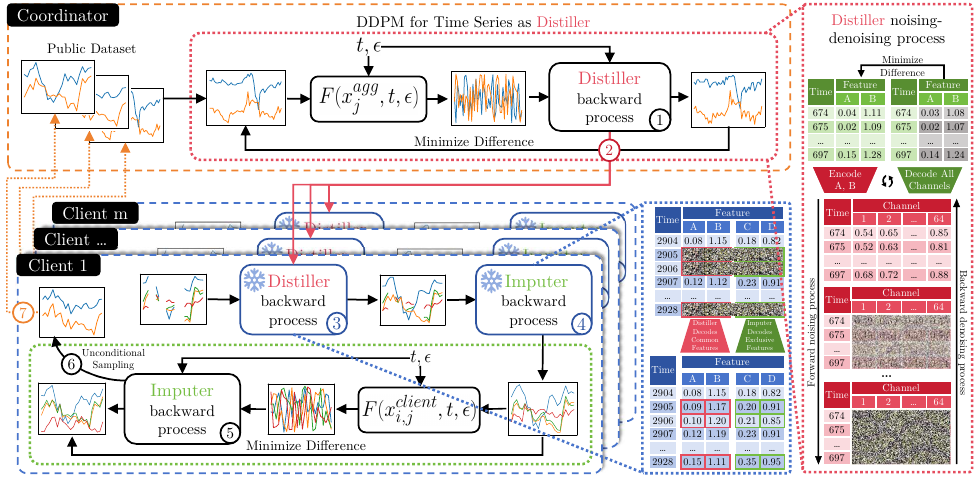}
    \caption{\sys Structure. First, the Distiller is pre-trained on a public dataset. Then, each client uses the distiller and imputer to impute common and exclusive features, respectively. Finally, synthetic data is sent back to the coordinator to expand the public dataset for the next round. The order of execution(1-7) is labeled in the figure. Here the common features are A and B, and the exclusive features are C and D.
    }
    \label{fig:Structure}
\end{figure}
 
In this work, we address the problem of collaborative time series imputation in the presence of temporal and feature misalignments, without requiring the sharing of raw data. In a federated learning setting, clients may possess different subsets of features. We categorize features into two types: \textit{common features} and \textit{exclusive features}. Common features are those present in all clients and also available in a public dataset, while exclusive features are unique to each client and not shared. For example, market indices might be common features in financial data, while individual portfolio holdings are exclusive. Our proposed framework, \sys, as shown in \autoref{fig:Structure}, tackles this problem using two models. A global distiller first imputes missing common features across clients. Local imputer models then use the imputed common features to predict the missing exclusive features for each client, addressing both temporal and feature misalignments. Furthermore, clients protect their privacy by sharing only synthetic versions of the common features while collaboratively improving the global distiller. This cycle of iterative imputation and model refinement ultimately converges to yield good quality imputations, while ensuring that no raw data is shared.

\subsection{Problem Definition}

We consider a federated learning setup involving $N$ clients and a coordinator. Each client $i$ possesses a time series dataset, denoted as \( \mathbf{X}^i = \left[ { X^i_{j,k} } \right]_{\{j = 1 \dots T^i, k = 1 \dots C^i\}} \), where $T^i$ is the number of time steps, and $C^i$ is the number of channels. These datasets can be split into two components, one for the common features and one for the exclusive features, i.e., $\mathbf{X}^i = \mathbf{X}^i_{\text{comm}} \cup \mathbf{X}^i_{\text{ex}}$. The coordinator holds a public dataset $\mathbf{X}^{\text{pub}} = \left[{ {X}^{\text{pub}}_{j,k} } \right]_{\forall j; k \in \mathcal{F}_{\text{comm}}}$, which contains data for the common features $\mathcal{F}_{\text{comm}}$ but without any missing values. This public dataset is time-indexed differently from the clients' data and provides a reliable reference for the common features. Each client's time series data comes from a distinct time interval, meaning that each client's time indices $j$ are unique. The feature set for each client $i$, $\mathcal{F}^i$, consists of common features $\mathcal{F}_{\text{comm}}$, which are shared across all clients, and exclusive features $\mathcal{F}^i_{\text{ex}}$, which are specific to each client. Thus, the overall feature set for client $i$ is represented as $\mathcal{F}^i = \mathcal{F}_{\text{comm}} \cup \mathcal{F}^i_{\text{ex}}$. Conversely, clients may have missing values in both the common and exclusive features. These missing values are indicated by a binary mask matrix $\mathbf{M}^i = \left[{ M^i_{j,k} }\right]_{\forall j,k}$, where $M^i_{j,k} = 1$ if the value $X^i_{j,k}$ is observed while $0$ indicates it is missing. The mask can be split into two parts: $\mathbf{M}^i_{\text{comm}}$, which corresponds to missing data in the common features, and $\mathbf{M}^i_{\text{ex}}$, which corresponds to missing data in the exclusive features. The goal is to design a collaborative method that enables clients to leverage shared knowledge and the public dataset to input the missing data locally without sharing raw data. Appendix A summarizes the mathematical notations used.

\subsection{Hybrid Federated Learning for Imputation Under Misalignment}
Algorithm~\ref{alg:FedTDD} presents the overview of \sys. The framework consists of two key components: the global distiller model $\mathcal{D}$ and the local imputer models $\mathcal{U}^i$. The global distiller $\mathcal{D}$ imputes missing common features shared across all clients, while each client trains a local imputer $\mathcal{U}^i$ to infer missing exclusive features specific to their data. These components work together to address temporal and feature misalignment by iteratively improving the imputation process over several rounds $r$ ranging from 1 to $R$.

\input{src/pseudocode/FedTDD_simple}

The process begins with the coordinator training a global distiller model $\mathcal{D}$ using the public dataset $\mathbf{X}^{\text{pub}}$. $\mathcal{D}$ leverages a temporal DDPM backbone to apply a forward diffusion process by gradually adding noise to the data and learns to reverse this process. During training, $\mathcal{D}$ conducts \emph{unconditional generation} by starting from Gaussian noise \( \boldsymbol{\epsilon} \) and learning to approximate the data distribution through the time and frequency domain components~\cite{yuan2024diffusion}. Formally, we have
\begin{equation}
    \mathcal{L}_{\text{time}} = \mathbb{E}_{(j,k,t) \, | \, \mathbf{M}^{\text{pub}}_{j,k} = 1} \left[ \left\| \mathbf{X}^{\text{pub}}_{j,k} - \Tilde{\mathbf{X}}^{\text{pub}}_{j,k}(\mathbf{X}_{j,k,t}^{\text{pub}}, t, \boldsymbol{\epsilon}; \theta) \right\|^2 \right] \quad \text{and}
\end{equation}
\begin{equation}
    \mathcal{L}_{\text{freq}} = \mathbb{E}_{(j,k,t) \, | \, \mathbf{M}^{\text{pub}}_{j,k} = 1} \left[ \left\| \text{FFT}(\mathbf{X}^{\text{pub}}_{j,k}) - \text{FFT}\left( \Tilde{\mathbf{X}}^{\text{pub}}_{j,k}(\mathbf{X}_{j,k,t}^{\text{pub}}, t, \boldsymbol{\epsilon}; \theta) \right) \right\|^2 \right],
\end{equation}
where \( \boldsymbol{\epsilon} \sim \mathcal{N}(0, \mathbf{I}) \), $\mathbf{X}^{\text{pub}}_{j,k}$ is the $(j,k)$-th entry of $\mathbf{X}^{\text{pub}}$, $\Tilde{\mathbf{X}}^{\text{pub}}_{j,k}$ is the denoised estimate from $\mathcal{D}$, and \(\text{FFT}\) denotes the Fast Fourier Transform~\cite{heckbert1995fourier}, which is a mathematical operation that converts a finite-length time domain signal to its frequency domain representation. We take the following objective
\begin{equation}
    \mathcal{L}_{\text{distiller}(\mathcal{D}^i)} = \mathbb{E}_{(j,k,t) \, | \, \mathbf{M}^{\text{pub}}_{j,k} = 1} \left[ w_t \left( \lambda_1 \mathcal{L}_{\text{time}} + \lambda_2 \mathcal{L}_{\text{freq}} \right) \right], \quad
    w_t = \frac{\lambda \gamma_t (1 - \bar{\gamma}_t)}{\delta_t^2},
\end{equation}
where $\lambda_1$ and $\lambda_2$ control the balance between time and frequency losses while $w_t$ emphasizes learning at larger diffusion steps, with $\lambda$ being a small constant. The parameter $\delta_t \in (0, 1)$ determines the amount of noise added at each forward diffusion step, where $t$ is a diffusion time step uniformly sampled from $1$ to $T$ during training.  The cumulative product \( \bar{\gamma}_t = \prod_{v=1}^t \gamma_v \), with \( \gamma_t = 1 - \delta_t \), track how the original signal diminishes over time due to the added noise. By weighting the loss at different steps, $w_t$ helps the model focus on reconstructing the signal under high-noise conditions. 

After this initial training, the coordinator distributes the trained global distiller model $\mathcal{D}$ to all participating clients. Each client $i$ then utilizes $\mathcal{D}$ to impute their missing common features. Since clients may have missing values in $\mathbf{X}_{\text{comm}}^i$, they input their data along with the corresponding mask $\mathbf{M}_{\text{comm}}^i$ to the distiller model, which will perform \emph{conditional generation} to iteratively refine the imputed data by sampling from the conditional distribution guided by the observed data, shown in Equation 17 in Appendix B.2.
The imputation process follows $\hat{\mathbf{X}}_{\text{comm}}^i = \mathcal{D}(\mathbf{X}_{\text{comm}}^i, \mathbf{M}_{\text{comm}}^i)$, where $\mathcal{D}$ reconstructs only the missing values, indicated by $\mathbf{M}_{\text{comm}}^i=0$. Similarly, the local imputer imputes missing values in $\mathbf{X}_{\text{ex}}^i$ by inputting their data along with the corresponding mask $\mathbf{M}_{\text{ex}}^i$ to the imputer model via $\hat{\mathbf{X}}_{\text{ex}}^i = \mathcal{U}(\mathbf{X}_{\text{ex}}^i, \mathbf{M}_{\text{ex}}^i)$. The imputed common features $\hat{\mathbf{X}}_{\text{comm}}^i$ are then combined with the available exclusive features $\hat{\mathbf{X}}_{\text{ex}}^i$ to form the training data $\mathbf{X}^i_{\text{train}} = \hat{\mathbf{X}}_{\text{comm}}^i \cup \hat{\mathbf{X}}_{\text{ex}}^i$ for the local imputer. Meanwhile, each client trains their local imputer model $\mathcal{U}^i$ using $\mathbf{X}^i_{\text{train}}$ as the ground truth. Since the imputed common features $\hat{\mathbf{X}}_{\text{comm}}^i$ are fully known (as they are outputs from the pre-trained and fine-tuned $\mathcal{D}$), they are entirely used as ground truth for training $\mathcal{U}^i$, regardless of the original mask $\mathbf{M}_{\text{comm}}^i$. For the exclusive features, only the observed entries indicated by the mask $\mathbf{M}_{\text{ex}}^i$ are used as ground truth since the quality of the imputer's generated data during training is not sufficient to be used as ground truth. We define the loss mask as $\mathbf{M}^i_{\text{loss}} = \mathbf{1}_{\text{comm}}^i \cup \mathbf{M}_{\text{ex}}^i$, where $\mathbf{1}_{\text{comm}}^i$ is a matrix of ones corresponding to the common features of client $i$. This loss mask ensures that the reconstruction loss is computed over all entries of the imputed common features and the observed entries of the exclusive features. The training loss for the imputer $\mathcal{U}^i$ can be defined as follows:
\begin{equation}
    \mathcal{L}_{\text{imputer}(\mathcal{U}^i)} = \mathbb{E}_{(j,k,t) \, | \, \mathbf{M}^i_{\text{loss}_{j,k}} = 1} \left[ w_t \left( \lambda_1 \mathcal{L}_{\text{time}}^i + \lambda_2 \mathcal{L}_{\text{freq}}^i \right) \right],
\end{equation}
\begin{equation*}
    \text{where} \quad \mathcal{L}_{\text{time}}^i = \mathbb{E}_{(j,k,t) \, | \, \mathbf{M}^i_{\text{loss}_{j,k}} = 1} \left[ \left\| \mathbf{X}^i_{\text{train}_{j,k}} - \Tilde{\mathbf{X}}^i_{\text{train}_{j,k}}(\mathbf{X}^i_{\text{train}_{j,k,t}}, t; \theta) \right\|^2 \right]
\end{equation*}
\begin{equation*}
    \text{and} \quad \mathcal{L}_{\text{freq}}^i = \mathbb{E}_{(j,k,t) \, | \, \mathbf{M}^i_{\text{loss}_{j,k}} = 1} \left[ \left\| \text{FFT}(\mathbf{X}^i_{\text{train}_{j,k}}) - \text{FFT}\left( \Tilde{\mathbf{X}}^i_{\text{train}_{j,k}}(\mathbf{X}^i_{\text{train}_{j,k,t}}, t; \theta) \right) \right\|^2 \right],
\end{equation*}
where $ \mathbf{X}^i_{\text{train}_{j,k}}$ is the $(j,k)$-th entry of $\mathbf{X}^i_{\text{train}}$, $\Tilde{\mathbf{X}}^i_{\text{train}_{j,k}}$ is the denoised estimate from $\mathcal{U}$. After training, each client uses the trained imputer $\mathcal{U}^i$ to generate a synthetic dataset through \emph{unconditional synthesis}, which includes both the common features $\hat{\mathbf{X}}_{\text{comm}}^i$ and the exclusive features $\hat{\mathbf{X}}_{\text{ex}}^i$. Starting from Gaussian noise, the imputer generates samples $\hat{\mathbf{X}}^i = \mathcal{U}^i(\mathbf{z}), \mathbf{z} \sim \mathcal{N}(0, \mathbf{I})$, that capture the distribution of both common and exclusive features.

To protect privacy, only the common features from the synthetic dataset, $\hat{\mathbf{X}}^i_{\text{comm}}$, are shared with the coordinator. This ensures that no raw or exclusive client data is exposed during the collaborative learning. The coordinator uses the synthetic common feature data from the clients to expand its public dataset. Rather than simply absorbing all the synthetic data, the coordinator carefully controls the growth of the dataset by accepting a fraction of the sequences from each client. Specifically, the coordinator adds $\frac{r}{R} \alpha * L$, where $L$ represents the length of the synthetic datasets $\hat{\mathbf{X}}^i_{\text{comm}}; \forall i \in \{1,2, \dots, N\}$, $\alpha$ is a hyperparameter between 0 and 1, and the ratio of $r$ and $R$ yields a number that linearly increases up to 1, allowing for a gradual expansion as the rounds increase. The coordinator retrains the global distiller $\mathcal{D}$ using this expanded dataset. The addition of synthetic data enhances the distiller's ability to learn the patterns necessary for imputing missing common features.

The overall process creates an iterative cycle of improvement. As clients' generative models, specifically their local imputers, become more accurate with each round, the quality of the synthetic data they generate also improves. This higher-quality synthetic data, in turn, improves the distiller model at the coordinator, which benefits all clients when it is redistributed. Over several training rounds, this mutual reinforcement drives both the global distiller and the local imputers to improve continuously. Ultimately, the process converges, yielding robust imputation models without requiring clients to share their raw data.

%% file: src/pseudocode/FedTDD_simple.tex
\SetCommentSty{textnormal} 
\SetKwComment{tcp}{$\triangleright$\ }{}

\begin{figure}
    \begin{algorithm}[H]
    \label{alg:FedTDD}
        \caption{\sys}
        \KwIn{Public dataset $\mathbf{X}^{\text{pub}}$, clients' datasets $\mathbf{X}^i$}
        \KwResult{Global distiller model $\mathcal{D}$, local imputer models $\mathcal{U}^i$}
        \textbf{Initialize:} Train $\mathcal{D}$ on $\mathbf{X}^{\text{pub}}$ 
        
        \For{$r = 1$ to $R$}{
        
            \For{each client $i$}{
            
                \textbf{Receive} global distiller $\mathcal{D}$ 
                
                $\hat{\mathbf{X}}^i_{\text{comm}} \leftarrow \mathcal{D}\left(\mathbf{X}_{\text{comm}}^i,\, \mathbf{M}_{\text{comm}}^i\right)$ \tcp*{\hspace{-0cm} Impute common features} 

                $\hat{\mathbf{X}}^i_{\text{ex}} \leftarrow \mathcal{U}\left(\mathbf{X}_{\text{ex}}^i,\, \mathbf{M}_{\text{ex}}^i\right)$ \tcp*{\hspace{-0cm} Impute exclusive features} 
                
                $\mathbf{X}^i_{\text{train}} \leftarrow \hat{\mathbf{X}}^i_{\text{comm}}\, \cup\, \hat{\mathbf{X}}^i_{\text{ex}}$ \tcp*{\hspace{-0cm} Combine with exclusive features} 
                
                \textbf{Train} $\mathcal{U}^i$ on $\mathbf{X}^i_{\text{train}}$  \tcp*{\hspace{-0cm} Train local imputer} 
                
                $\hat{\mathbf{X}}^i \leftarrow \mathcal{U}^i(\mathbf{z}), \mathbf{z} \sim \mathcal{N}(0, \mathbf{I})$ \tcp*{\hspace{-0cm} Generate synthetic data} 
                
                \textbf{Send} $\hat{\mathbf{X}}^i_{\text{comm}}$ from  $\hat{\mathbf{X}}^i$ to coordinator
            }
            
            \For{each client $i$}{
            
                Select $n_r = \dfrac{r}{R}\, \alpha \cdot L$ sequences from $\hat{\mathbf{X}}_{\text{comm}}^i$ 
                
                
                $\mathbf{X}^{\text{pub}} \leftarrow \mathbf{X}^{\text{pub}}\, \cup\, \hat{\mathbf{X}}_{\text{comm}}^i[1:n_r]$ \tcp*{\hspace{-0cm} Expand public dataset} 
            }
            \textbf{Finetune} $\mathcal{D}$ on updated $\mathbf{X}^{\text{pub}}$
        }
    \end{algorithm}
\end{figure}

%% file: src/experiments.tex
\section{Experiments}

We assess \sys 's performance by showing its advantages and disadvantages when applied to multiple benchmark datasets. We leave the analysis of different training configurations in the Appendix C, where we examine the impact of limited public data, abundant sequences with missing data, imbalanced data distributions and different aggregation strategies on model performance.

\paragraph{Datasets} To assess the quality of synthetic data, we consider four real-world datasets and one simulated dataset with different properties, such as the number of features, correlation, periodicity, and noise levels. Each dataset is preprocessed using a sliding window technique~\cite{yoon2019time} to segment the data into sequences of length 24 to capture meaningful temporal dependencies while keeping the computational cost manageable. \textbf{Stocks} is the daily historical Google stock data from 2004 to 2019 with highly correlated features. \textbf{ETTh} recorded the electricity transformers hourly between July 2016 and July 2018, including load and oil temperature data that consists of 7 features. \textbf{Energy} from UCI appliances energy prediction dataset with 10-minute intervals for about 4.5 months. \textbf{fMRI} is a realistic simulation of brain activity time series with 50 features. \textbf{MuJoCo} is a physics-based simulation time series containing 14 features. We show the statistics of all datasets in Appendix D.2.

\begin{figure}
    \centering
    \begin{subfigure}[t]{0.3\linewidth}
        \centering
        \frame{\includegraphics[width=1\textwidth]{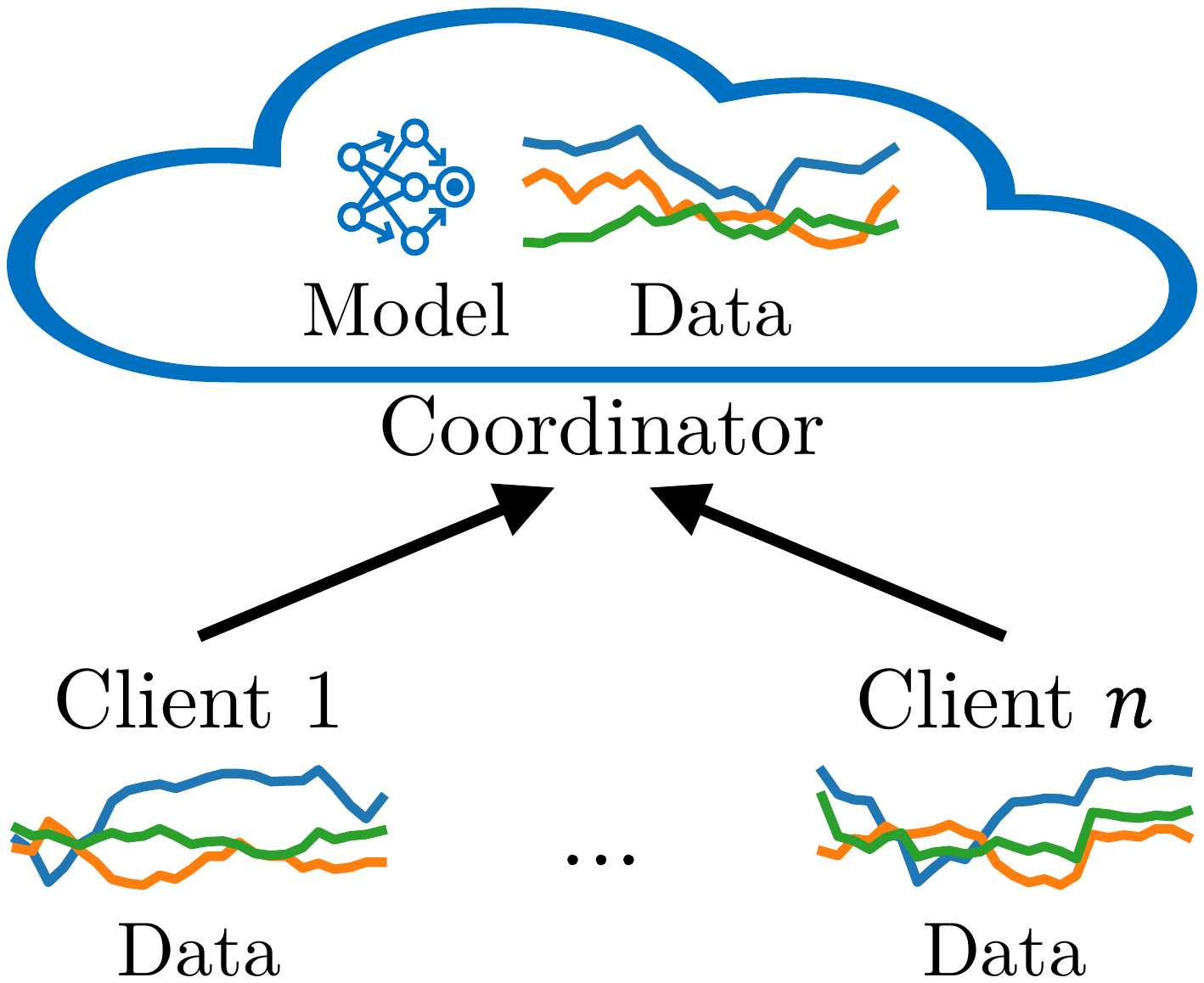}}
        \caption{Centralized*}
        \label{fig:baselines_centralized_star}
    \end{subfigure}
    \hfill
    \begin{subfigure}[t]{0.3\linewidth}
        \centering
        \frame{\includegraphics[width=1\textwidth]{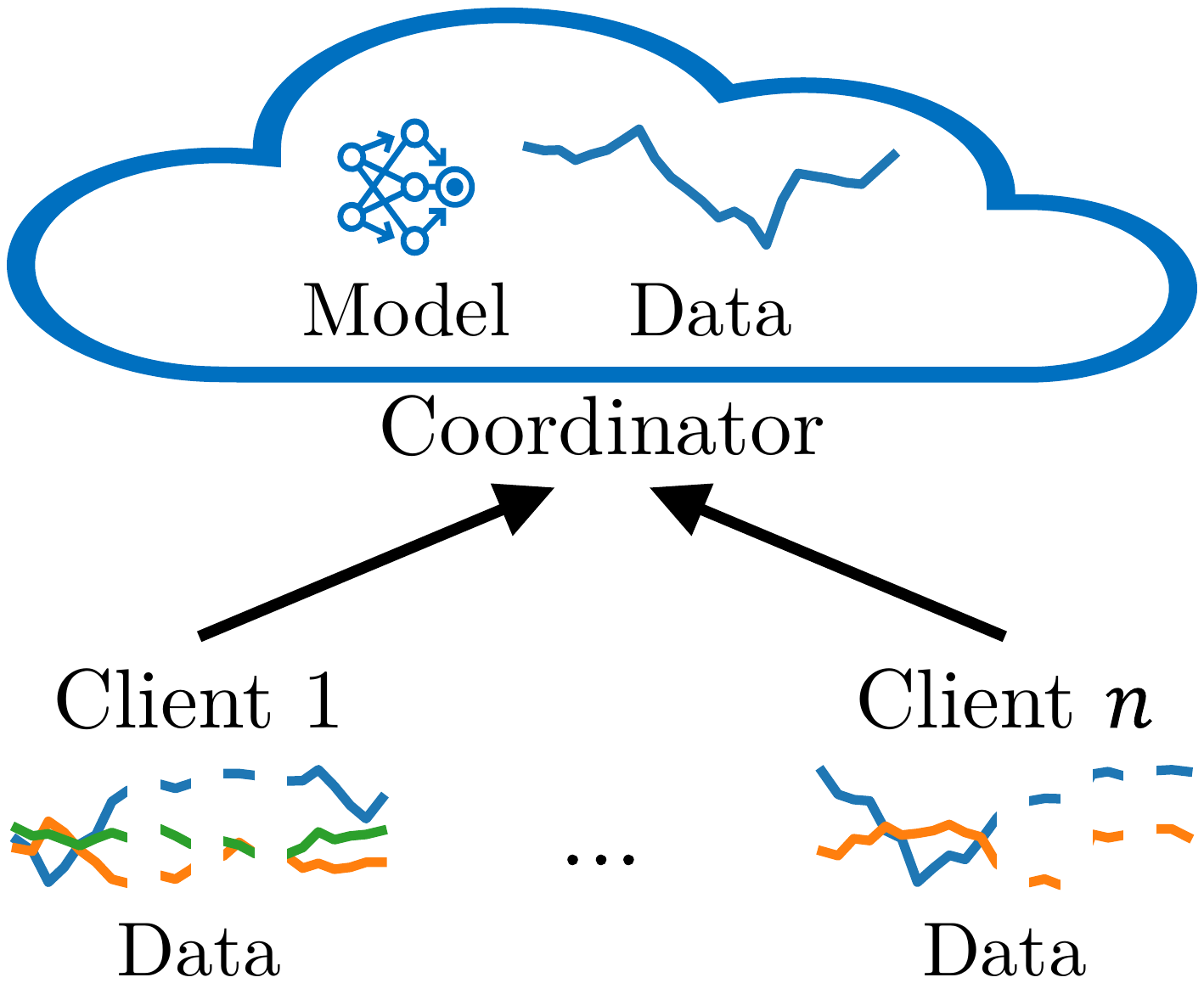}}
        \caption{Centralized}
        \label{fig:baselines_centralized}
    \end{subfigure}
    \hfill
    \begin{subfigure}[t]{0.3\linewidth}
        \centering
        \frame{\includegraphics[width=1\textwidth]{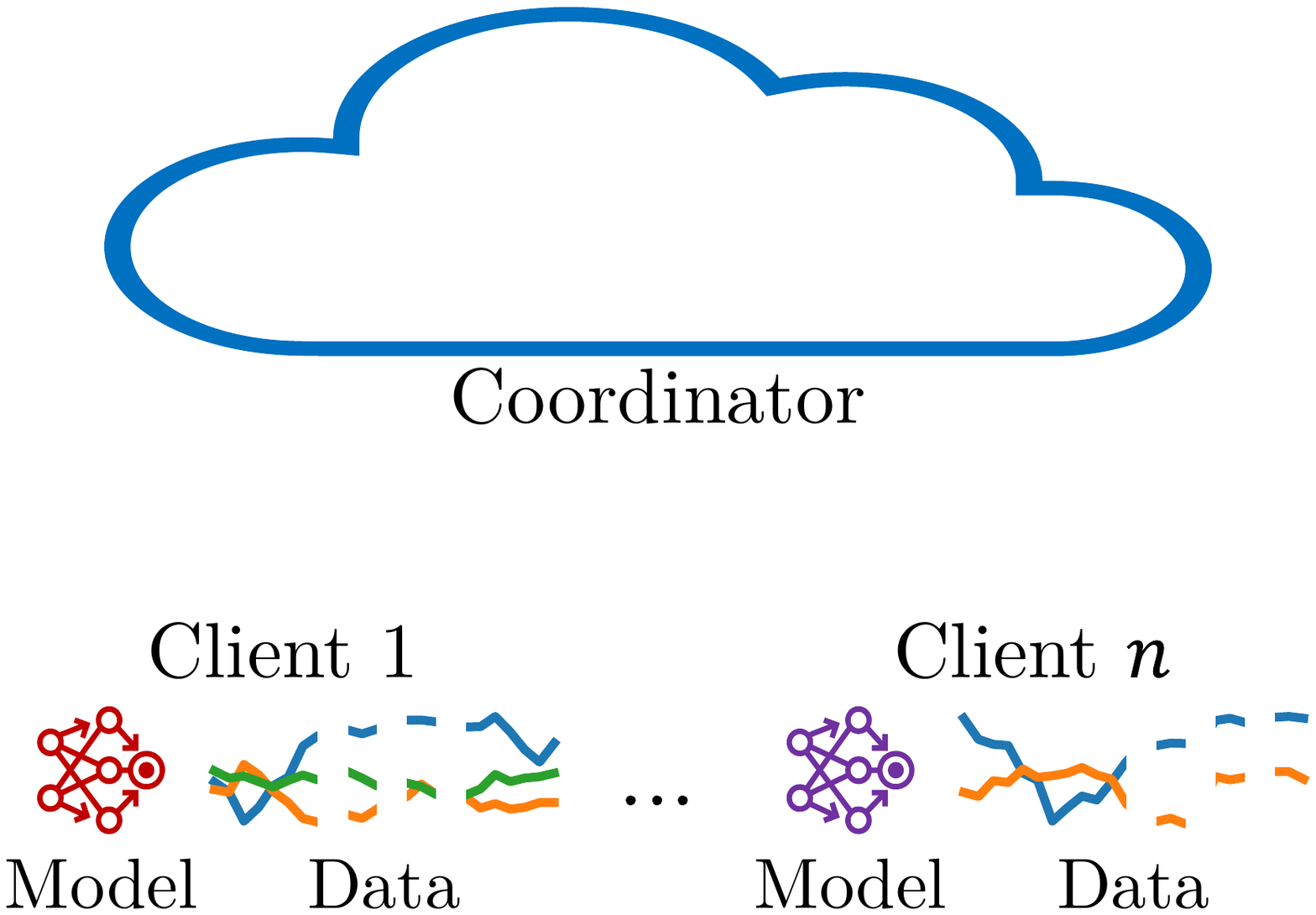}}
        \caption{Local}
        \label{fig:baselines_local}
    \end{subfigure}
    \vskip 0.1in
    \hspace*{\fill}
    \begin{subfigure}[t]{0.3\linewidth}
        \centering
        \frame{\includegraphics[width=1\textwidth]{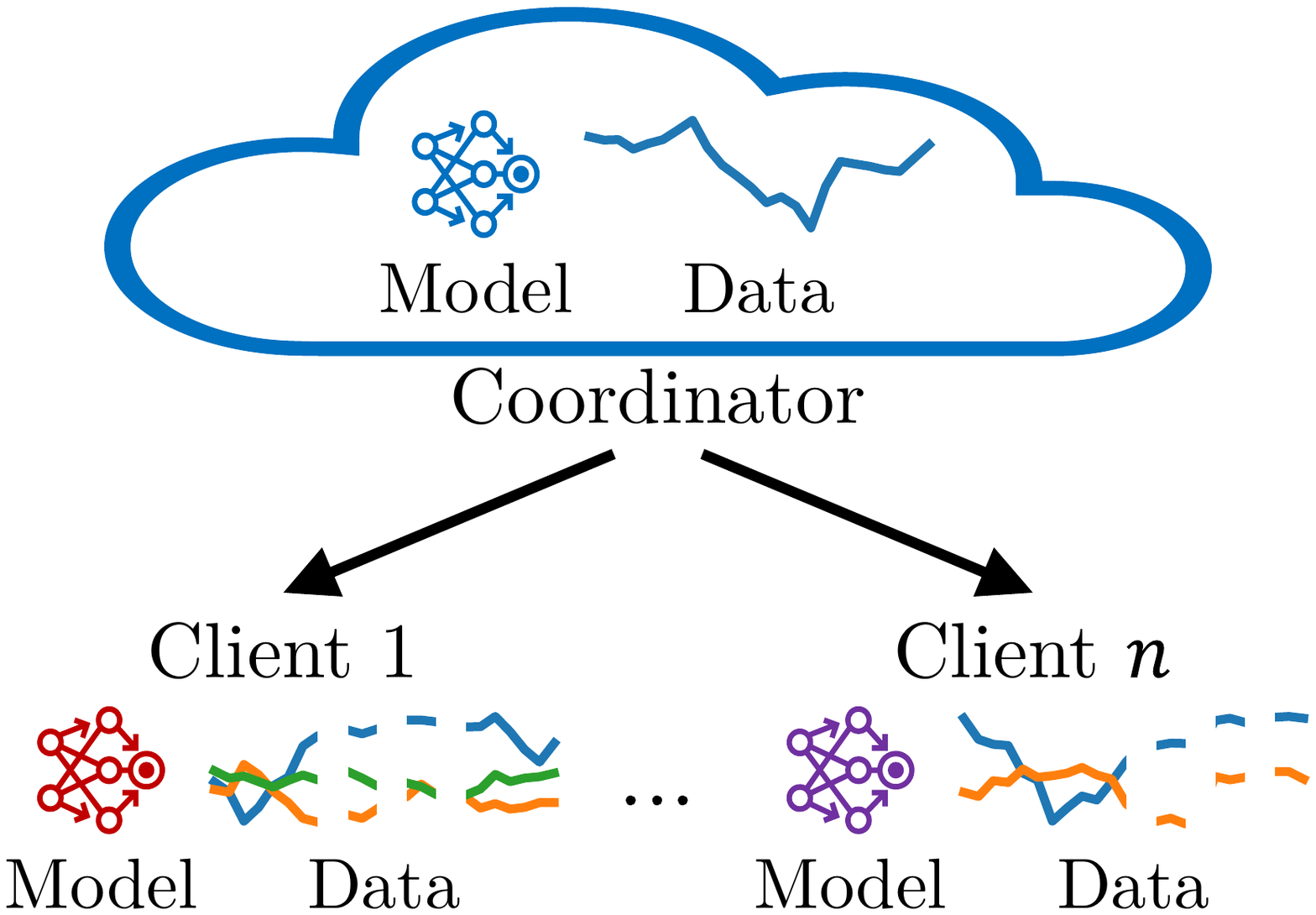}}
        \caption{Pre-trained}
        \label{fig:baselines_pretrained}
    \end{subfigure}
    \hfill
    \begin{subfigure}[t]{0.3\linewidth}
        \centering
        \frame{\includegraphics[width=1\textwidth]{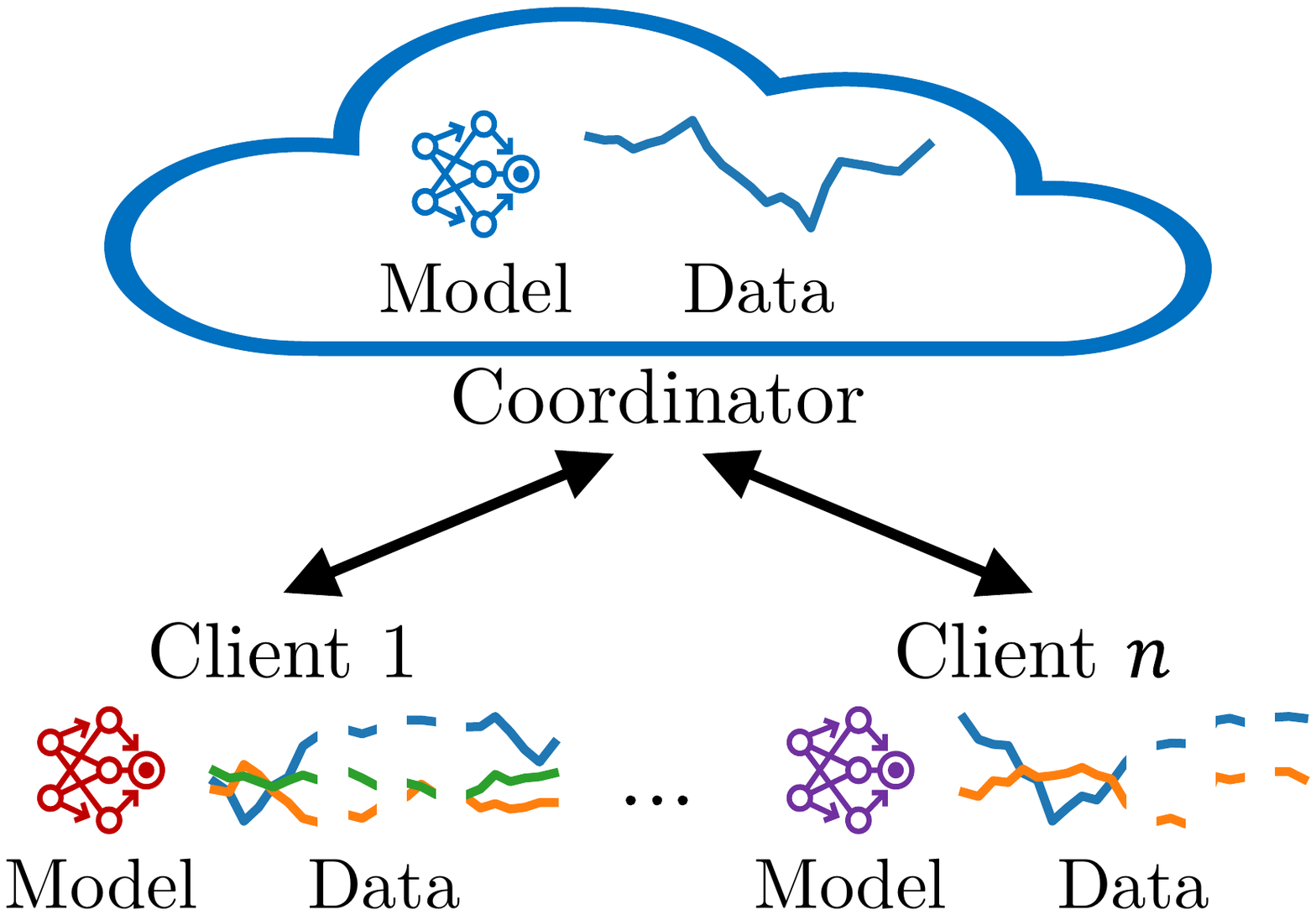}}
        \caption{\sys}
        \label{fig:baselines_fedtdd}
    \end{subfigure}
    \hspace*{\fill}
    \caption{Illustrations of different baselines compared to \sys. The data in the coordinator, also called public data, in \autoref{fig:baselines_centralized}, \ref{fig:baselines_pretrained} and \ref{fig:baselines_fedtdd} consists only common features time series. Dashes indicate temporal missing values.}
    \label{fig:baselines}
\end{figure}

\paragraph{Baselines} We compare \sys against approaches show in \autoref{fig:baselines_centralized_star}, \ref{fig:baselines_centralized}, \ref{fig:baselines_local} and \ref{fig:baselines_pretrained}. For the \textbf{Centralized*} training, we aggregate all data from individual clients, including public data, into a single location, where a global model is trained using the combined dataset, and this will be trained with all available features in the dataset and without missing values. While \textbf{Centralized} uses the same training procedure as Centralized*, it is, however, trained on a combined dataset with missing values and corresponding features available from each client plus the public data. To deal with differing features across clients, we create the combined dataset consisting of the total number of features in the particular benchmark dataset and zero-fill any remaining features to ensure uniformity. On the other hand, \textbf{Local} training involves training a separate model for each client using only their local data, without any communication or data aggregation. This approach has to be done to verify that \sys can perform relatively better than train locally. Finally, the \textbf{Pre-trained} approach leverages a model trained on a public dataset and uses it to impute the common features in local data from each client. Again, there is no data aggregation for this approach.
In comparison, \sys integrates the Pre-trained approach and applies data aggregation to it. We utilized a SOTA diffusion-based multivariate time series generative model, Diffusion-TS~\cite{yuan2024diffusion}, as the backbone for these baselines and \sys. Alternatively, any other time series generative model can be adopted in these approaches in a plug-and-play manner.

\paragraph{Evaluation metrics} We quantitatively assess the quality of the generated synthetic data using four key metrics (see Appendix D.3 for more details). \textbf{Context-Fréchet Inception Distance (Context-FID) score}~\cite{jeha2022psa} evaluates the similarity between the distribution of real and synthetic time series data by computing the Fréchet distance. \textbf{Correlational score}~\cite{liao2020conditional} measures the correlation between the features of multivariate time series in the synthetic data compared to its real data. \textbf{Discriminative score}~\cite{yoon2019time} measures the realism of the synthetic data by training a binary classifier to distinguish between real and synthetic data. \textbf{Predictive score}~\cite{yoon2019time} evaluates the utility of the synthetic data by training a sequence-to-sequence model on the synthetic data and measuring its performance on real data. All evaluation metrics are computed based on the respective features of the individual clients and then averaged over five trials, followed by calculating the overall average across the number of clients. The quality of synthetic data is considered the ``best'' when all metrics approach 0, meaning lower values indicate better quality.

\paragraph{Training configurations} We run \sys and the baselines mentioned above with ten clients, five global rounds, 7500 local epochs for the first round, and 5000 for the rest. Besides, the coordinator trains on the public data consisting of common features, and each client contributes a set of features, which is the combination of common and exclusive features. The number of common features is around 50\% of the total number of features in the original dataset. On the other hand, we use \ac{pr} to manipulate the proportion of the public data that has to be reserved from the entire dataset before partitioning the dataset to all clients. \Ac{sr} divides all sequences into two groups. In the first group, a mask is applied to just the common features, while in the second group, the mask is applied to all features.
Moreover, \ac{mr} is the missing rate to mask on a sequence of multivariate time series, and we consider the missing scenario as shown in Appendix D.4. In the main experiments, we set \ac{pr}, \ac{sr}, and \ac{mr} to 0.5. All the hyperparameters are listed in Appendix D.5.
\label{sec:training_config}


\subsection{Time Series Generation}

\begin{figure}
    \centering
    \includegraphics[width=1\textwidth]{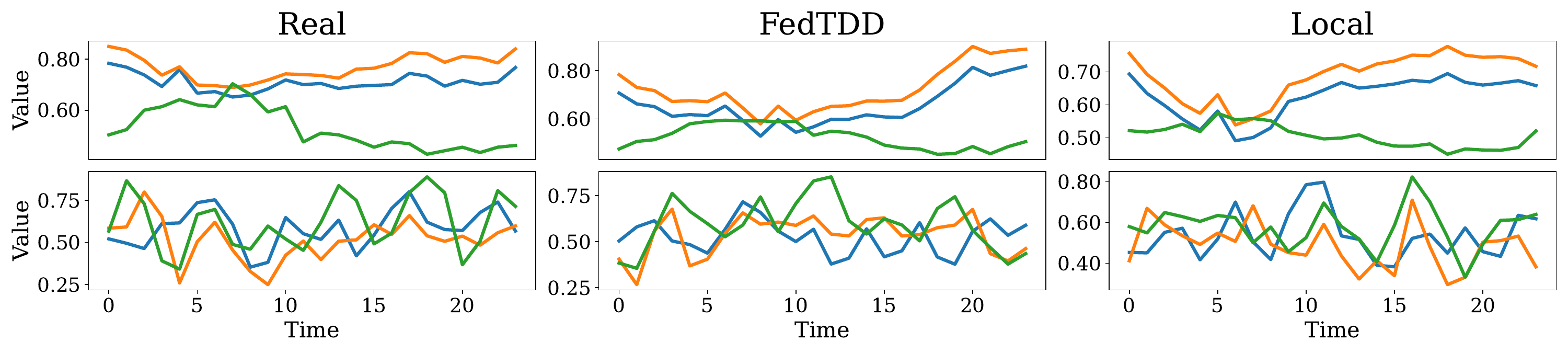}
    \caption{Real samples and synthetic samples generated unconditionally from \sys and Local. The first and second rows of samples are from ETTh and fMRI datasets, respectively.}
    \label{fig:main_ts}
\end{figure}

In \autoref{tab:main_exp}, we quantitatively analyze the quality of unconditionally generated 24-length time series for diverse time series datasets. \sys shows a strong performance comparable to the Centralized* approach. The proposed aggregation mechanism during fine-tuning proved essential to prevent the degradation of the coordinator model's performance and, in turn, the client models. By doing this, we achieved strong results across most datasets. We also present the generated synthetic samples of one representative client for ETTh and fMRI datasets in \autoref{fig:main_ts}.

\input{src/tables/main_exp}

\paragraph{Challenges on fMRI dataset} We observe that the fMRI dataset's imputation quality was lower than other datasets, as the mean square error between the imputed and real data is greater. Consequently, client models degraded due to training on low-quality imputed data. This suggests that the imputation strategy may need further refinement for such datasets, where the data distribution and complexity present greater challenges for accurate synthetic data generation and imputation. Besides, the Local approach achieves the best Correlational and Discriminative scores for the fMRI dataset. However, we cannot conclude that training locally is the best overall approach for fMRI. As we mentioned, the low performance of \sys and Pre-trained is primarily due to the poor quality of the imputed data, which affects training. This shows the advantage of Local training not relying on imputed data, making it seem better suited for the fMRI dataset compared to \sys and Pre-trained.
\label{sec:challenges_fmri}

\paragraph{Comparison between Centralized and Local training} 
Both Centralized and Local approaches are trained on datasets with missing values, but their performance differs significantly. This could be due to the different model architectures used in each approach. As aforementioned, the Centralized model is trained on a combined dataset where the additional features are filled with zeros, which results in the worst performance. This shows the advantage of having an individual model trained locally for each client.

\subsection{Ablation Study}

In \autoref{tab:main_ablation}, we show the result of reducing the number of common features in \sys. We set the number of common features to around 25\% of the total number of features in the corresponding dataset. As a result, we can observe the robustness of \sys when dealing with a relatively small number of common features across most datasets. However, \sys does not perform as expected on the fMRI dataset because of the poor quality of imputed data, as mentioned in Section~\ref{sec:challenges_fmri}. On the other hand, the performance of Centralized training slightly decreased due to more zeros filling out the combined dataset, especially in the public data.

\input{src/tables/main_ablation}

%% file: src/tables/main_exp.tex
\begin{table}
    \centering
    \captionsetup{width=1\textwidth}
    \caption{Results on multiple time series datasets. \textbf{Bold} indicates best performance.}
    \label{tab:main_exp}
    \resizebox{\textwidth}{!}{
    \begin{tabular}{l l c c c c c}
        \toprule[1.5pt]
        \multirow{1}{*}{\bf{Metric}}
        & \multirow{1}{*}{\bf{Method}}
        & \multirow{1}{*}{\bf{Stocks}}
        & \multirow{1}{*}{\bf{ETTh}}
        & \multirow{1}{*}{\bf{MuJoCo}}
        & \multirow{1}{*}{\bf{Energy}}
        & \multirow{1}{*}{\bf{fMRI}} \\
        \midrule
        \multirow{5}{*}{Context-FID}
        & Centralized*  & 0.682\scriptsize{$\pm$0.106}       & 0.281\scriptsize{$\pm$0.040}       & 0.782\scriptsize{$\pm$0.138}       & 0.533\scriptsize{$\pm$0.082}       & 1.737\scriptsize{$\pm$0.125}       \\
        & Centralized   & 3.548\scriptsize{$\pm$0.990}       & 8.870\scriptsize{$\pm$2.295}       & 10.00\scriptsize{$\pm$2.814}       & 9.343\scriptsize{$\pm$2.808}       & 13.56\scriptsize{$\pm$3.357}       \\
        & Local         & 1.648\scriptsize{$\pm$0.229}       & 1.313\scriptsize{$\pm$0.188}       & 0.751\scriptsize{$\pm$0.121}       & 1.179\scriptsize{$\pm$0.179}       & 1.694\scriptsize{$\pm$0.153}       \\
        & Pre-trained   & 1.047\scriptsize{$\pm$0.169}       & 0.326\scriptsize{$\pm$0.040}       & 0.617\scriptsize{$\pm$0.090}       & 0.412\scriptsize{$\pm$0.054}       & \bf{1.411\scriptsize{$\pm$0.102}}  \\
        & \sys          & \bf{0.675\scriptsize{$\pm$0.087}}  & \bf{0.271\scriptsize{$\pm$0.038}}  & \bf{0.529\scriptsize{$\pm$0.068}}  & \bf{0.376\scriptsize{$\pm$0.056}}  & 1.459\scriptsize{$\pm$0.099}       \\
        \midrule
        \multirow{5}{*}{Correlational}
        & Centralized*  & 0.061\scriptsize{$\pm$0.043}       & 0.253\scriptsize{$\pm$0.094}       & 1.989\scriptsize{$\pm$0.247}       & 5.231\scriptsize{$\pm$1.294}       & 7.900\scriptsize{$\pm$0.384}       \\
        & Centralized   & 0.769\scriptsize{$\pm$0.336}       & 0.340\scriptsize{$\pm$0.097}       & 2.230\scriptsize{$\pm$0.518}       & 5.681\scriptsize{$\pm$0.634}       & 18.07\scriptsize{$\pm$2.311}       \\
        & Local         & 0.156\scriptsize{$\pm$0.120}       & 0.239\scriptsize{$\pm$0.079}       & 1.298\scriptsize{$\pm$0.260}       & 3.447\scriptsize{$\pm$0.838}       & \bf{5.992\scriptsize{$\pm$0.383}}  \\
        & Pre-trained   & 0.077\scriptsize{$\pm$0.052}       & 0.165\scriptsize{$\pm$0.074}       & 1.323\scriptsize{$\pm$0.171}       & 2.821\scriptsize{$\pm$0.651}       & 6.049\scriptsize{$\pm$0.349}       \\
        & \sys          & \bf{0.058\scriptsize{$\pm$0.050}}  & \bf{0.161\scriptsize{$\pm$0.064}}  & \bf{1.296\scriptsize{$\pm$0.215}}  & \bf{2.800\scriptsize{$\pm$0.686}}  & 6.017\scriptsize{$\pm$0.364}       \\
        \midrule
        \multirow{5}{*}{Discriminative}
        & Centralized*  & 0.136\scriptsize{$\pm$0.091}       & 0.199\scriptsize{$\pm$0.061}       & 0.297\scriptsize{$\pm$0.108}       & 0.230\scriptsize{$\pm$0.080}       & 0.422\scriptsize{$\pm$0.074}       \\
        & Centralized   & 0.476\scriptsize{$\pm$0.042}       & 0.475\scriptsize{$\pm$0.017}       & 0.474\scriptsize{$\pm$0.024}       & 0.496\scriptsize{$\pm$0.006}       & 0.477\scriptsize{$\pm$0.030}       \\
        & Local         & 0.340\scriptsize{$\pm$0.153}       & 0.298\scriptsize{$\pm$0.060}       & 0.200\scriptsize{$\pm$0.092}       & 0.329\scriptsize{$\pm$0.087}       & \bf{0.397\scriptsize{$\pm$0.061}}  \\
        & Pre-trained   & \bf{0.175\scriptsize{$\pm$0.117}}  & 0.115\scriptsize{$\pm$0.060}       & 0.208\scriptsize{$\pm$0.068}       & \bf{0.141\scriptsize{$\pm$0.068}}  & 0.419\scriptsize{$\pm$0.051}       \\
        & \sys          & 0.185\scriptsize{$\pm$0.105}       & \bf{0.106\scriptsize{$\pm$0.061}}  & \bf{0.153\scriptsize{$\pm$0.120}}  & 0.153\scriptsize{$\pm$0.072}       & 0.414\scriptsize{$\pm$0.051}       \\
        \midrule
        \multirow{5}{*}{Predictive}
        & Centralized*  & 0.040\scriptsize{$\pm$0.000}       & 0.127\scriptsize{$\pm$0.003}       & 0.112\scriptsize{$\pm$0.015}       & 0.292\scriptsize{$\pm$0.009}       & 0.137\scriptsize{$\pm$0.004}       \\
        & Centralized   & 0.047\scriptsize{$\pm$0.012}       & 0.223\scriptsize{$\pm$0.020}       & 0.165\scriptsize{$\pm$0.060}       & 0.427\scriptsize{$\pm$0.053}       & 0.233\scriptsize{$\pm$0.051}       \\
        & Local         & 0.043\scriptsize{$\pm$0.003}       & 0.118\scriptsize{$\pm$0.011}       & 0.048\scriptsize{$\pm$0.006}       & 0.204\scriptsize{$\pm$0.012}       & 0.135\scriptsize{$\pm$0.006}       \\
        & Pre-trained   & 0.046\scriptsize{$\pm$0.001}       & 0.104\scriptsize{$\pm$0.004}       & 0.052\scriptsize{$\pm$0.004}       & 0.177\scriptsize{$\pm$0.005}       & 0.133\scriptsize{$\pm$0.006}       \\
        & \sys          & \bf{0.041\scriptsize{$\pm$0.001}}  & \bf{0.101\scriptsize{$\pm$0.004}}  & \bf{0.048\scriptsize{$\pm$0.004}}  & \bf{0.175\scriptsize{$\pm$0.006}}  & \bf{0.133\scriptsize{$\pm$0.004}}  \\
        \bottomrule[1.5pt]
    \end{tabular}}
\end{table}

%% file: src/tables/main_ablation.tex
\begin{table}
    \centering
    \captionsetup{width=\linewidth}
    \caption{Ablation study for a relatively small number of common features. \textbf{Bold} indicates best performance.}
    \label{tab:main_ablation}
    \resizebox{\linewidth}{!}{
    \begin{tabular}{l l c c c c c}
        \toprule[1.5pt]
        \multirow{1}{*}{\bf{Metric}}
        & \multirow{1}{*}{\bf{Method}}
        & \multirow{1}{*}{\bf{Stocks}}
        & \multirow{1}{*}{\bf{ETTh}}
        & \multirow{1}{*}{\bf{MuJoCo}}
        & \multirow{1}{*}{\bf{Energy}}
        & \multirow{1}{*}{\bf{fMRI}} \\
        \midrule
        \multirow{5}{*}{Context-FID}
        & Centralized*  & 0.682\scriptsize{$\pm$0.106}       & 0.281\scriptsize{$\pm$0.040}       & 0.782\scriptsize{$\pm$0.138}       & 0.533\scriptsize{$\pm$0.082}       & 1.737\scriptsize{$\pm$0.125}       \\
        & Centralized   & 3.733\scriptsize{$\pm$0.959}       & 11.54\scriptsize{$\pm$3.894}       & 14.68\scriptsize{$\pm$4.263}       & 13.17\scriptsize{$\pm$3.035}       & 15.34\scriptsize{$\pm$4.789}       \\
        & Local         & 1.982\scriptsize{$\pm$0.234}       & 0.824\scriptsize{$\pm$0.105}       & 0.660\scriptsize{$\pm$0.100}       & 0.844\scriptsize{$\pm$0.127}       & 1.220\scriptsize{$\pm$0.098}       \\
        & Pre-trained   & 0.738\scriptsize{$\pm$0.142}       & 0.316\scriptsize{$\pm$0.032}       & 0.547\scriptsize{$\pm$0.099}       & 0.381\scriptsize{$\pm$0.066}       & \bf{1.178\scriptsize{$\pm$0.104}}  \\
        & \sys          & \bf{0.680\scriptsize{$\pm$0.123}}  & \bf{0.267\scriptsize{$\pm$0.036}}  & \bf{0.510\scriptsize{$\pm$0.072}}  & \bf{0.331\scriptsize{$\pm$0.051}}  & 1.196\scriptsize{$\pm$0.098}       \\
        \midrule
        \multirow{5}{*}{Correlational}
        & Centralized*  & 0.061\scriptsize{$\pm$0.043}       & 0.253\scriptsize{$\pm$0.094}       & 1.989\scriptsize{$\pm$0.247}       & 5.231\scriptsize{$\pm$1.294}       & 7.900\scriptsize{$\pm$0.384}       \\
        & Centralized   & 0.697\scriptsize{$\pm$0.168}       & 0.523\scriptsize{$\pm$0.095}       & 2.317\scriptsize{$\pm$0.597}       & 5.781\scriptsize{$\pm$0.924}       & 31.35\scriptsize{$\pm$4.923}       \\
        & Local         & 0.091\scriptsize{$\pm$0.052}       & 0.167\scriptsize{$\pm$0.057}       & 1.079\scriptsize{$\pm$0.196}       & 1.984\scriptsize{$\pm$0.594}       & \bf{4.929\scriptsize{$\pm$0.395}}  \\
        & Pre-trained   & 0.028\scriptsize{$\pm$0.027}       & \bf{0.132\scriptsize{$\pm$0.054}}  & 1.115\scriptsize{$\pm$0.233}       & 1.795\scriptsize{$\pm$0.577}       & 5.033\scriptsize{$\pm$0.323}       \\
        & \sys          & \bf{0.025\scriptsize{$\pm$0.022}}  & 0.137\scriptsize{$\pm$0.064}       & \bf{1.060\scriptsize{$\pm$0.209}}  & \bf{1.737\scriptsize{$\pm$0.282}}  & 5.005\scriptsize{$\pm$0.317}       \\
        \midrule
        \multirow{5}{*}{Discriminative}
        & Centralized*  & 0.136\scriptsize{$\pm$0.091}       & 0.199\scriptsize{$\pm$0.061}       & 0.297\scriptsize{$\pm$0.108}       & 0.230\scriptsize{$\pm$0.080}       & 0.422\scriptsize{$\pm$0.074}       \\
        & Centralized   & 0.475\scriptsize{$\pm$0.041}       & 0.469\scriptsize{$\pm$0.020}       & 0.479\scriptsize{$\pm$0.026}       & 0.494\scriptsize{$\pm$0.010}       & 0.484\scriptsize{$\pm$0.023}       \\
        & Local         & 0.300\scriptsize{$\pm$0.116}       & 0.208\scriptsize{$\pm$0.070}       & 0.190\scriptsize{$\pm$0.088}       & 0.241\scriptsize{$\pm$0.071}       & \bf{0.398\scriptsize{$\pm$0.058}}  \\
        & Pre-trained   & 0.119\scriptsize{$\pm$0.088}       & 0.116\scriptsize{$\pm$0.067}       & 0.163\scriptsize{$\pm$0.088}       & 0.130\scriptsize{$\pm$0.058}       & 0.418\scriptsize{$\pm$0.050}       \\
        & \sys          & \bf{0.112\scriptsize{$\pm$0.097}}  & \bf{0.107\scriptsize{$\pm$0.078}}  & \bf{0.157\scriptsize{$\pm$0.104}}  & \bf{0.120\scriptsize{$\pm$0.067}}  & 0.412\scriptsize{$\pm$0.057}       \\
        \midrule
        \multirow{5}{*}{Predictive}
        & Centralized*  & 0.040\scriptsize{$\pm$0.000}       & 0.127\scriptsize{$\pm$0.003}       & 0.112\scriptsize{$\pm$0.015}       & 0.292\scriptsize{$\pm$0.009}       & 0.137\scriptsize{$\pm$0.004}       \\
        & Centralized   & 0.168\scriptsize{$\pm$0.025}       & 0.196\scriptsize{$\pm$0.027}       & 0.198\scriptsize{$\pm$0.049}       & 0.314\scriptsize{$\pm$0.052}       & 0.223\scriptsize{$\pm$0.029}       \\
        & Local         & 0.084\scriptsize{$\pm$0.038}       & 0.114\scriptsize{$\pm$0.009}       & 0.069\scriptsize{$\pm$0.010}       & 0.199\scriptsize{$\pm$0.007}       & \bf{0.130\scriptsize{$\pm$0.005}}  \\
        & Pre-trained   & 0.028\scriptsize{$\pm$0.007}       & 0.108\scriptsize{$\pm$0.004}       & 0.063\scriptsize{$\pm$0.007}       & 0.190\scriptsize{$\pm$0.005}       & 0.132\scriptsize{$\pm$0.005}       \\
        & \sys          & \bf{0.028\scriptsize{$\pm$0.005}}  & \bf{0.107\scriptsize{$\pm$0.005}}  & \bf{0.062\scriptsize{$\pm$0.006}}  & \bf{0.186\scriptsize{$\pm$0.004}}  & \bf{0.130\scriptsize{$\pm$0.005}}  \\
        \bottomrule[1.5pt]
    \end{tabular}}
\end{table}

%% file: src/conclusion.tex
\section{Conclusion}

While federated learning is increasingly applied for different regression tasks for time series (TS), it is still limited in handling generative tasks, especially when time series features are vertically partitioned and temporarily misaligned. We propose a novel federated TS generation framework, \sys, which trains TS diffusion model by leveraging the self-imputing capability of the diffusion model and globally aggregating from clients' knowledge through data distillation and clients' synthetic data. The central component of \sys is a distiller at the coordinator that first is pre-trained on the public datasets and then periodically fine-tuned by the aggregated intermediate synthetic data from the clients. Clients keep their personalized TS diffusion models and train them with local data and synthetic data of the latest distiller periodically. Our extensive evaluation across five datasets shows that \sys effectively overcomes the hurdle of feature partition and temporal misalignment, achieving improvements of up to 79.4\% and 62.8\% over local training on Context-FID and Correlational scores, while delivering performance comparable to centralized baselines.

\begin{credits}
\subsubsection{\ackname} This research is part of the Priv-GSyn, 200021E\_229204, of Swiss National Science Foundation and the DEPMAT project, P20-22 / N21022, of the research programme Perspectief which is partly financed by the Dutch Research Council (NWO).

\subsubsection{\discintname}
The authors have no competing interests to declare that are relevant to the content of this article.
\end{credits}




%% file: src/bibliography.tex
\bibliographystyle{splncs04}
\bibliography{src/refs}

%% file: main.bbl
\begin{thebibliography}{10}
\providecommand{\url}[1]{\texttt{#1}}
\providecommand{\urlprefix}{URL }
\providecommand{\doi}[1]{https://doi.org/#1}

\bibitem{alaa2021generative}
Alaa, A., Chan, A.J., van~der Schaar, M.: Generative time-series modeling with fourier flows. In: International Conference on Learning Representations (2021)

\bibitem{alcaraz2022diffusion}
Alcaraz, J.M.L., Strodthoff, N.: Diffusion-based time series imputation and forecasting with structured state space models. arXiv preprint arXiv:2208.09399  (2022)

\bibitem{brophy2021estimation}
Brophy, E., De~Vos, M., Boylan, G., Ward, T.: Estimation of continuous blood pressure from ppg via a federated learning approach. Sensors  \textbf{21}(18), ~6311 (2021)

\bibitem{ceneda2018guided}
Ceneda, D., Gschwandtner, T., Miksch, S., Tominski, C.: Guided visual exploration of cyclical patterns in time-series. In: Proceedings of the IEEE Symposium on Visualization in Data Science (VDS). IEEE Computer Society (2018)

\bibitem{desai2021timevae}
Desai, A., Freeman, C., Wang, Z., Beaver, I.: Timevae: A variational auto-encoder for multivariate time series generation. arXiv preprint arXiv:2111.08095  (2021)

\bibitem{fons2022hypertime}
Fons, E., Sztrajman, A., El-Laham, Y., Iosifidis, A., Vyetrenko, S.: Hypertime: Implicit neural representation for time series. arXiv preprint arXiv:2208.05836  (2022)

\bibitem{goodfellow2020generative}
Goodfellow, I., Pouget-Abadie, J., Mirza, M., Xu, B., Warde-Farley, D., Ozair, S., Courville, A., Bengio, Y.: Generative adversarial networks. Communications of the ACM  \textbf{63}(11),  139--144 (2020)

\bibitem{heckbert1995fourier}
Heckbert, P.: Fourier transforms and the fast fourier transform (fft) algorithm. Computer Graphics  \textbf{2}(1995),  15--463 (1995)

\bibitem{ho2020denoising}
Ho, J., Jain, A., Abbeel, P.: Denoising diffusion probabilistic models. Advances in neural information processing systems  \textbf{33},  6840--6851 (2020)

\bibitem{jeha2022psa}
Jeha, P., Bohlke-Schneider, M., Mercado, P., Kapoor, S., Nirwan, R.S., Flunkert, V., Gasthaus, J., Januschowski, T.: Psa-gan: Progressive self attention gans for synthetic time series. In: The Tenth International Conference on Learning Representations (2022)

\bibitem{kingma2013auto}
Kingma, D.P.: Auto-encoding variational bayes. arXiv preprint arXiv:1312.6114  (2013)

\bibitem{kitagawa1984smoothness}
Kitagawa, G., Gersch, W.: A smoothness priors--state space modeling of time series with trend and seasonality. Journal of the American Statistical Association  \textbf{79}(386),  378--389 (1984)

\bibitem{kollovieh2024predict}
Kollovieh, M., Ansari, A.F., Bohlke-Schneider, M., Zschiegner, J., Wang, H., Wang, Y.B.: Predict, refine, synthesize: Self-guiding diffusion models for probabilistic time series forecasting. Advances in Neural Information Processing Systems  \textbf{36} (2024)

\bibitem{li2020federated}
Li, T., Sahu, A.K., Zaheer, M., Sanjabi, M., Talwalkar, A., Smith, V.: Federated optimization in heterogeneous networks. Proceedings of Machine learning and systems  \textbf{2},  429--450 (2020)

\bibitem{liao2020conditional}
Liao, S., Ni, H., Szpruch, L., Wiese, M., Sabate-Vidales, M., Xiao, B.: Conditional sig-wasserstein gans for time series generation. arXiv preprint arXiv:2006.05421  (2020)

\bibitem{lim2021time}
Lim, B., Zohren, S.: Time-series forecasting with deep learning: a survey. Philosophical Transactions of the Royal Society A  \textbf{379}(2194),  20200209 (2021)

\bibitem{liu2024vertical}
Liu, Y., Kang, Y., Zou, T., Pu, Y., He, Y., Ye, X., Ouyang, Y., Zhang, Y.Q., Yang, Q.: Vertical federated learning: Concepts, advances, and challenges. IEEE Transactions on Knowledge and Data Engineering  (2024)

\bibitem{luu2021time}
Luu, K., Khashabi, D., Gururangan, S., Mandyam, K., Smith, N.A.: Time waits for no one! analysis and challenges of temporal misalignment. arXiv preprint arXiv:2111.07408  (2021)

\bibitem{mcmahan2017communication}
McMahan, B., Moore, E., Ramage, D., Hampson, S., y~Arcas, B.A.: Communication-efficient learning of deep networks from decentralized data. In: Artificial intelligence and statistics. pp. 1273--1282. PMLR (2017)

\bibitem{meijer2025ts}
Meijer, C., Huang, J., Sharma, S., Lazovik, E., Chen, L.Y.: Ts-inverse: A gradient inversion attack tailored for federated time series forecasting models. In: 2025 IEEE Conference on Secure and Trustworthy Machine Learning (SaTML). pp. 110--124. IEEE (2025)

\bibitem{mendieta2022local}
Mendieta, M., Yang, T., Wang, P., Lee, M., Ding, Z., Chen, C.: Local learning matters: Rethinking data heterogeneity in federated learning. In: Proceedings of the IEEE/CVF Conference on Computer Vision and Pattern Recognition. pp. 8397--8406 (2022)

\bibitem{mogren2016c}
Mogren, O.: C-rnn-gan: Continuous recurrent neural networks with adversarial training. arXiv preprint arXiv:1611.09904  (2016)

\bibitem{pratama2016review}
Pratama, I., Permanasari, A.E., Ardiyanto, I., Indrayani, R.: A review of missing values handling methods on time-series data. In: 2016 international conference on information technology systems and innovation (ICITSI). pp.~1--6. IEEE (2016)

\bibitem{qu2022rethinking}
Qu, L., Zhou, Y., Liang, P.P., Xia, Y., Wang, F., Adeli, E., Fei-Fei, L., Rubin, D.: Rethinking architecture design for tackling data heterogeneity in federated learning. In: Proceedings of the IEEE/CVF conference on computer vision and pattern recognition. pp. 10061--10071 (2022)

\bibitem{rasouli2020fedgan}
Rasouli, M., Sun, T., Rajagopal, R.: Fedgan: Federated generative adversarial networks for distributed data. arXiv preprint arXiv:2006.07228  (2020)

\bibitem{rasul2021autoregressive}
Rasul, K., Seward, C., Schuster, I., Vollgraf, R.: Autoregressive denoising diffusion models for multivariate probabilistic time series forecasting. In: International Conference on Machine Learning. pp. 8857--8868. PMLR (2021)

\bibitem{sachdeva2023data}
Sachdeva, N., McAuley, J.: Data distillation: A survey. arXiv preprint arXiv:2301.04272  (2023)

\bibitem{shankar2024silofuse}
Shankar, A., Brouwer, H., Hai, R., Chen, L.: Silofuse: Cross-silo synthetic data generation with latent tabular diffusion models (2024)

\bibitem{song2021federated}
Song, J., Ye, J.C.: Federated cyclegan for privacy-preserving image-to-image translation. arXiv preprint arXiv:2106.09246  (2021)

\bibitem{tashiro2021csdi}
Tashiro, Y., Song, J., Song, Y., Ermon, S.: Csdi: Conditional score-based diffusion models for probabilistic time series imputation. Advances in Neural Information Processing Systems  \textbf{34},  24804--24816 (2021)

\bibitem{vaswani2017attention}
Vaswani, A.: Attention is all you need. Advances in Neural Information Processing Systems  (2017)

\bibitem{gdpr}
Voigt, P., Von~dem Bussche, A.: The eu general data protection regulation (gdpr). A Practical Guide, 1st Ed., Cham: Springer International Publishing  \textbf{10}(3152676),  10--5555 (2017)

\bibitem{wang2023differentially}
Wang, Z., Cheng, X., Su, S., Wang, G.: Differentially private generative decomposed adversarial network for vertically partitioned data sharing. Information Sciences  \textbf{619},  722--744 (2023)

\bibitem{yoon2019time}
Yoon, J., Jarrett, D., Van~der Schaar, M.: Time-series generative adversarial networks. Advances in neural information processing systems  \textbf{32} (2019)

\bibitem{yuan2024diffusion}
Yuan, X., Qiao, Y.: Diffusion-ts: Interpretable diffusion for general time series generation. arXiv preprint arXiv:2403.01742  (2024)

\bibitem{yuan2024vflgan}
Yuan, X., Zhao, Z., Gope, P., Sikdar, B.: Vflgan-ts: Vertical federated learning-based generative adversarial networks for publication of vertically partitioned time-series data. arXiv preprint arXiv:2409.03612  (2024)

\bibitem{zhang2021survey}
Zhang, C., Xie, Y., Bai, H., Yu, B., Li, W., Gao, Y.: A survey on federated learning. Knowledge-Based Systems  \textbf{216},  106775 (2021)

\bibitem{zhao2023gtv}
Zhao, Z., Wu, H., Van~Moorsel, A., Chen, L.Y.: Gtv: generating tabular data via vertical federated learning. arXiv preprint arXiv:2302.01706  (2023)

\bibitem{zhu2017unpaired}
Zhu, J.Y., Park, T., Isola, P., Efros, A.A.: Unpaired image-to-image translation using cycle-consistent adversarial networks. In: Proceedings of the IEEE international conference on computer vision. pp. 2223--2232 (2017)

\end{thebibliography}
